\documentclass[11pt]{article}
\usepackage{geometry}
\usepackage{coling2020}
\usepackage{times}
\usepackage{url}
\usepackage{latexsym}
\usepackage{microtype}
\usepackage{hyperref}
\usepackage{amsmath}
\usepackage{verbatim}
\usepackage{xargs}                      

\usepackage{multirow}
\usepackage{multicol}
\usepackage{rotating}
\usepackage[disable]{todonotes}

\newcommand\stil[1]{\todo[author=Steve,color=purple!40,inline]{#1}}

\newcommand\ajil[1]{\todo[author=Anupam,color=green!40,inline]{#1}}

\newcommandx{\unsure}[2][1=]{\todo[linecolor=red,backgroundcolor=red!25,bordercolor=red,#1]{#2}}
\newcommandx{\change}[2][1=]{\todo[linecolor=blue,backgroundcolor=blue!25,bordercolor=blue,#1]{#2}}
\newcommandx{\info}[2][1=]{\todo[linecolor=OliveGreen,backgroundcolor=OliveGreen!25,bordercolor=OliveGreen,#1]{#2}}
\newcommandx{\improvement}[2][1=]{\todo[linecolor=pink,backgroundcolor=pink!25,bordercolor=pink,#1]{#2}}
\newcommandx{\thiswillnotshow}[2][1=]{\todo[disable,#1]{#2}}

\newcommand\tabletop{\hline\noalign{\smallskip}}
\newcommand\tablemid{\noalign{\smallskip}\hline\noalign{\smallskip}}
\newcommand\tablebot{\noalign{\smallskip}\hline}

\colingfinalcopy 

\title{NIT-Agartala-NLP-Team at SemEval-2020 Task 8:\\ Building Multimodal Classifiers to tackle Internet Humor}

\author{ 
Steve Durairaj Swamy \hspace*{0.5em} Shubham Laddha  \hspace*{0.5em} Basil Abdussalam 
\AND
Debayan Datta \hspace*{0.5em} Anupam Jamatia  \\
Department of  Computer Science and Engineering \\
National Institute of Technology \\
Agartala, Tripura, India \\
{\tt \small{\{steve050798, laddha.shubham97, basilwkrh, debayan.datta98, anupamjamatia\}@gmail.com
}}}

\date{}

\begin{document}
\maketitle
\begin{abstract}
  The paper describes the systems submitted to SemEval-2020 Task 8: Memotion by the `NIT-Agartala-NLP-Team'. A dataset of 8879 memes was made available by the task organizers to train and test our models. Our systems include a Logistic Regression baseline, a BiLSTM + Attention-based learner and a transfer learning approach with BERT. For the three sub-tasks A, B and C, we attained ranks 24/33, 11/29 and 15/26, respectively. We highlight our difficulties in harnessing image information as well as some techniques and handcrafted features we employ to overcome these issues. We also discuss various modelling issues and theorize possible solutions and reasons as to why these problems persist. 
\end{abstract}

\section{Introduction}
\label{sec:intro}

Over the years, the internet and social media have become an indispensable part of our lives. Today, an average netizen spends over 45 minutes on some form of social media everyday\footnote{\url{www.bit.ly/2NZQ5L5}}. Therefore, social media has become a goldmine for data to model and study human opinions and behaviour. Social media analysis conventionally deals with data in the form of text, audio or video --- but often only one prominent modality is studied in isolation. This lack of multimodality in research works has led to entire modes of communication being disregarded. One such form of internet communication that we have yet to tap into is memes. While a meme was initially defined as \textit{``an idea, behaviour, or style that spreads from person to person within a culture—often to convey a particular phenomenon, theme, or meaning\footnote{\url{www.lexico.com/en/definition/meme}},"} it has transformed into the umbrella term for a suite of referential humor that plagues the internet. Memes can more aptly be defined as \textit{``a form of referential humor that incites humor by leveraging images, text and sometimes audio."} The image is more often than not coupled with a real-life event or media. The images are then re-purposed to incite humor through the subversion of expectations and via reference. Through the years, memes have been leveraged differently by various communities --- apart from innocent, humor inducing purposes. Corporations and companies are increasingly interested in harnessing memes to sway their younger customers. More adversely, certain memes such as \textit{Pepe the Frog}\footnote{\url{https://en.wikipedia.org/wiki/Pepe_the_Frog}} were adopted by alt-right groups and used as a calling card of sorts.
Memes are also used as performative acts, which involve a conscious decision to support or oppose a movement (as seen in the case of the recent Hong Kong protests \footnote{\url{www.pri.org/stories/2019-07-16/memes-hong-kong-protests}}). The abundance of memes on social media platforms such as Facebook, Instagram, and Twitter \cite{SavvasEA:2018} further suggests that --- these digital constructs have become a flagship of internet culture, so and so that, to understand memes would mean to understand the views of a community. The SemEval Task 8: Memotion \cite{ChhaviEA2020} shared task is one such grassroots endeavour to understand memes better and incorporate multimodality in social media analysis. Apart from the rich information that could be derived from understanding social media, solving the various challenges of the task also pose unique research value --- such as the training of models that could understand reference derived from background knowledge and complex human expressions such as humor and sarcasm. 

In this paper, we share the insights we gain from our time working with the task and dataset. In Section ~\ref{sec:related}, we discuss some previous work and related approaches to tackling the problem. Then, in Section~\ref{sec:data}, we introduce the task and the dataset. Subsequently, in Section~\ref{sec:methodology}, we outline the features employed and the model architectures. Section~\ref{sec:results} reports our results and model performance. In Section~\ref{sec:error_analysis} and \ref{sec:discussion}, we analyze our results and discuss the possible source and solutions to our errors and issues. We summarize our key results and conjectures in Section \ref{sec:conclusion}.
\section{Related Work}
\label{sec:related}

Research on Memes, in particular, are few and far in between. Some works have experimented with memes as sentiment predictors: In their work on Facebook sentiment analysis, \newcite{French2017} reported a positive correlation between the category of the meme used and the affection of the discussion from its texts. \newcite{French2017} go on to confirm that memes were far more successful in conveying the sentiment of the debate over textual data. Other works attempt to automate the meme generation process \cite{Pierson2018,Oliveira2016} but fail at replicating the humor expressed by particular memes. While \newcite{Pierson2018} attempts to re-purpose image captioning algorithms to capture the humor, \newcite{Oliveira2016} attempt to use macros within news headlines to do the same. However, both works fail to generate coherent humor that can persuade annotators consistently. Another similar work by \newcite{WangEA2015} apply multimodal techniques to generate captions for popular meme formats.
Finally, the propagation and effect of memes on social media are also studied in works such as the one by \newcite{ZannettouEA2018} which provides an assessment of the popularity and use of certain memes, in the context of each community. In another work,  \newcite{FerraraEA2013} aim to detect and analyze meme usage in social media streams, particularly by using an unsupervised clustering framework. Work in the space also make use of meta-information \cite{WangEA2015,FerraraEA2013} regarding the meme to provide context to their frameworks. Apart from memes, other work on multimodal social media analysis and humor analysis could provide valuable insights. \newcite{ShinEA2018} show how multimodal information can be used to enhance the analysis of unstructured data on social media. There are numerous works that attempt to predict the sentiment of images \cite{Kanishcheva2015,XuEA2014}. However, these works focus on a single modality --- images. Works on Multimodal sentiment analysis like the one by \newcite{HuEA2018} use both image and text embeddings in their models to show slight improvements over the text-only models. A comprehensive overview of multimodal sentiment analysis can be found in work, such as that \newcite{Kaur2019} and  \newcite{Soleymani2017}. In the realm of humor analysis --- work such as the recent paper by \newcite{WellerSeppi2019} have shown that deep learning models can outperform humans in classifying humor mainly due to a disparity in the sense of humor of the annotators and the testers. In an interesting work by \newcite{ChandrasekaranEA2018} an attempt is made to capture wit by using synonyms of words in image descriptions to incite puns through the subversion of expectations. While the work reports impressive results and beats out humans in a controlled vocabulary setting, humans quickly regain dominance when the vocabulary is unconstrained.

\section{Data}
\label{sec:data}

The task organizers have made available a dataset \cite{ChhaviEA2020} of 8879
\ajil{then in somewhere in the document you have written that we have un- annotated test data} annotated memes scrapped from various sources across the internet. Each meme 
\stil{I have mentioned it in the experimental setup sir}
\ajil{i am confused about the annotated and un-annotated term!}
was annotated by two annotators to ensure annotation quality. The text was extracted from the image using the Google OCR system and manually corrected by crowdsourced workers --- to ensure that model accuracy doesn't depend on the quality of the OCR techniques used. We briefly describe each subtask of the Memotion Shared Task below:
 
\begin{itemize} 
    \item \textbf{Task A --- Sentiment Classification:} Given a meme, the task is to classify it as a positive, negative or neutral meme.
    \item \textbf{Task B --- Multilabel Characteristic Classification:} Given a meme, the system has to identify the existence of the following characteristics --- humor, sarcasm, offense and motivation. Being a multilabel classification task a meme can exhibit any combination of the above characteristics or none at all.
    \item \textbf{Task C --- Scales of Semantic Classes:} The third task is to quantify the extent to which a particular effect is being expressed, i.e. if the meme is humorous whether it is funny, very funny or hilarious and so on.
\end{itemize}
 
The dataset shows a significant imbalance, particularly in the  representation of the labels {\tt negative}, {\tt hilarious}, {\tt very\_twisted} and {\tt hateful\_offensive} of the sentiment, humor, sarcasm and offensive categories, respectively. We represent the distribution of labels in Table \ref{tab:dataset}.

\begin{table*}[ht!]
    \small
    \centering
    \addtolength{\tabcolsep}{1pt} 
    \begin{tabular}{l|l|c|c}
        \tabletop
        Categories & Tags & No. of Samples & Percentage(\%)\\
        \tablemid
        \multirow{3}{*}{Sentiment Analysis}
        & {\tt negative} & 631 & 09.01 \\
        & {\tt neutral} & 2,205 & 31.50 \\
        & {\tt positive} & 4165 & 59.49 \\
        \tablemid
        \multirow{4}{*}{Humor}
        & {\tt not\_funny} & 1,651 & 23.58 \\ 
        & {\tt funny} & 2,457 & 35.10 \\
        & {\tt very\_funny} & 2,241 & 32.01 \\
        & {\tt hilarious} & 652 & 09.31 \\
        \tablemid
        \multirow{4}{*}{Sarcasm}
        & {\tt general}  & 3,512 & 50.16 \\ 
        & {\tt not\_sarcastic} & 1,546 & 22.09 \\
        & {\tt twisted\_meaning} & 1,549 & 22.12 \\
        & {\tt very\_twisted} & 394 & 05.63 \\
        \tablemid
        \multirow{4}{*}{Offensive}
        & {\tt not\_offensive}  & 2,715 & 38.78 \\ 
        & {\tt slight\_offensive} & 2,596 &  37.08 \\
        & {\tt very\_offensive} & 1,469 & 20.98 \\
        & {\tt hateful\_offensive} & 221 & 03.16 \\
        \tablemid
        \multirow{2}{*}{Motivational}
        & {\tt not\_motivational}  & 4,530 & 64.70 \\ 
        & {\tt motivational} & 2,471 &  35.30 \\
        \tablebot
    \end{tabular}
    
    \addtolength{\tabcolsep}{-1pt}  
    \caption{Sample distribution of the corpus}
    \label{tab:dataset}
\end{table*}


\section{Preprocessing and System Overview}
\label{sec:methodology}

Before we outline our features and models, we digress to briefly explain the steps we undertook to reduce noise in the text. We perform basic lower casing and the removal of unnecessary punctuation as the initial step followed by more specific noise inducing aspects such as the removal of URLs and User mentions (using regular expressions).


We did not perform any global image preprocessing steps; however, we did perform basic preprocessing steps such as resizing and grayscale conversion for specific feature extraction techniques.

\subsection{Features} 
Depending on the model we use, we employ three different text vectorization techniques --- word (1,2)-gram {\tt TFIDF} features for the Logistic Regression model; A GLoVe \cite{PenningtonEA:2014} embedding pretrained on 27 billion tweets for the BiLSTM + Attention based model; Contextual {\tt BERT} embeddings \cite{DevlinEA2019} for our transfer learning approach.

Beyond, text vectorization we explore a few \emph{hand crafted features} to improve model performance. We use feature previously employed by \newcite{BeteroEA16} in their work to detect humor in sitcoms and \newcite{MahajanEA2017}'s submission to the SemEval 2017 Task 6. 
The stylistic features are as follows: number of words, number of parts-of-speech (POS) tags such as nouns, adjectives and verbs and their ratio to the number of words. The POS tags mentioned are obtained using CMU POS Tagger \cite{OlutobiEA:13}.

Ambiguity features are useful in representing the multiple meanings that can be delivered simultaneously as found in pun related humor \cite{YangEA:2015,Miller:2015}. 
For this purpose we use a concept called \textit{Synset} (short for Synonym set). A Synset is defined as a set of one or more synonyms that can be interchangeably used in the same context to express the same meaning which was originally embedded. We derive these synonyms using the NLTK Corpus \cite{LoperBird:09}. For example the Synset for the word `\textit{new}' would be \{ new, fresh, raw, newfangled, modern, newly \}. The ambiguity features used are as follows: Mean Synset Length (it is the mean of the length of synset of each word of the text)
, Maximum Synset Length (it is the maximum length of synset that a single word in the text can have), Synset Length Gap (it the difference between the Maximum Synset Length and Mean Synset Length).

Initially, we experimented with the use of pretrained feature extractors to extract image information. Our approach, using various pretrained models trained on the ILSVRC \cite{imagenet_cvpr09} such as the popular Inception\_V2\_ResNet \cite{SzegedyEA2017}, exhibited underwhelming results and sometimes proved to be detrimental to model performance. In this regard, we elected to employ more hand crafted features rather than pretrained feature extractors. For each image: hue, saturation and value are calculated by converting the RGB images into HSV channels using the scikit-image toolkit \cite{scikit-image}. We average the hue, saturation and value over all the pixels in the image to obtain the hue ($H_{image}$), saturation ($S_{image}$) and luminance ($V_{image}$) of the image. We also include RMS contrast features for each image --- which can simply be defined as the standard deviation of pixel intensities (i.e. the brightness).
We draw inspiration from previous work by \newcite{ZhangEA15} and explore more features that can be quantitatively derived from the HSV model such as Colourfulness --- a metric defined \newcite{HaslerEA03} that exhibits high correlation to human perception of colourfulness --- and metrics by \newcite{ValdezEA94} features -- that relate brightness and saturation to the following emotions: Pleasure, Arousal and Dominance.
Following this, we also entertain the idea of employing facial expressions as an extra image feature. To do this we re-purpose an opensource emotion detection application \footnote{\url{www.github.com/atulapra/Emotion-detection}}. The approach used a convolution neural network at its core, to detect the following emotions --- Angry, Disgusted, Fearful, Happy, Neutral, Sad and Surprised. The model is pretrained on the FER2013 Kaggle dataset \footnote{\url{www.kaggle.com/deadskull7/fer2013}} and used as feature extractor in our pipeline. 
At this juncture, it might be apt to recognise the small dataset size and label imbalance exhibited by the Memotion Dataset \cite{ChhaviEA2020}. To tackle the imabalance and data scarcity we consider many techniques such as oversampling, weighted training and resampling techniques (such as SMOTE \cite{ChawlaEA2002}) in our pipelines. 
We also consider a text augmentation technique found in work from \newcite{Zhang2015}. We re-purpose open source code found on github \footnote{\url{www.github.com/Opla/SmallData-Augmentation-MachineLearning}}. The core of the idea is to extend the given dataset by simple word replacement wherein we replace certain words in the sentence with corresponding synonyms or phrase replacements pulled from an auxiliary database/dictionary. 
The image features are then replicated for the newly created synthetic sample. Our database of choice for such replacements was the \textit{paraphrase bank}\footnote{\url{www.paraphrase.org}} database. We apply this technique on the training split of the data and duplicate the corresponding image and dense features.

\subsection{Model Descriptions}
For a baseline approach to the given problem, we elected an  L2 regularized Logistic Regression as an ideal starting point. The input representation for text was {\tt TF-IDF} uni-grams and bi-grams. We also use the handcrafted text and image dense features as well the emotion vectors with this model. All the modelling was done using the sci-kit learn toolkit \cite{PedregosaEA:11}. This model uses the previously explained data augmentation technique as well as balanced class weights during training.

\begin{figure}[t]
    \centering
    \includegraphics[scale = 0.32]{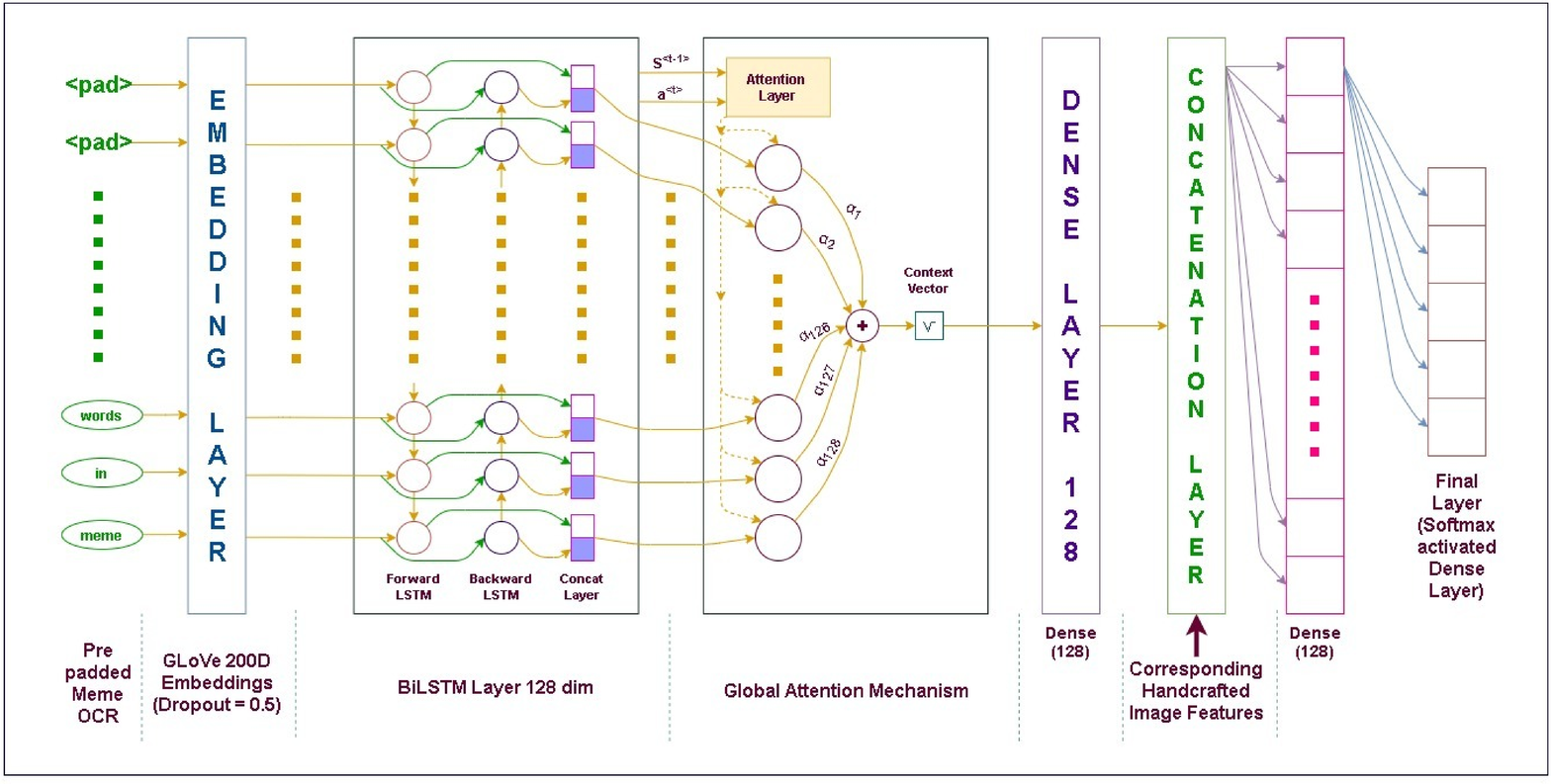}
    \caption{Attention informed model Architecture}
    \label{fig:Attention}
\end{figure}

Our second model is an attention based deep learning model. Attention is a technique first introduced in the machine translation research space by Bahdanau et. al. \cite{BahdanauEA14}. The idea was further extended to text classification by work such as the ones by \newcite{YangEA16} and \newcite{RaffelEA15} --- which are what we implement here. All modelling tasks for this model was carried out using keras with a tensorflow \cite{AbadiEA2015} backend. The model uses the 'Adam' learning rate optimizer and categorical cross entropy loss function during the training phase. The model architecture is more intuitively described in Figure \ref{fig:Attention}. Due to the small dataset size we associate large dropout values with the embedding layer (0.5), LSTM layers (0.3) and Dense layers (0.3, apart from a L2 regularization kernel). We also implement early stopping, to further alleviate overfitting. This model also makes use of the data augmentation technique previously mentioned and SMOTE \cite{ChawlaEA2002} for balancing/resampling purposes. The model was set to train for 100 epochs with a batch size of 2048 and initial learning rate of 0.001 but we found that the model started overfitting at an average of 40 epochs.

We also explore the possibility of leveraging dynamic contextual embeddings through the use of a transfer learning model --- {\tt BERT}. The developers of BERT provide a simple classification API for BERT through the {\tt run\_classifier} API available on their github page \footnote{\url{www.github.com/google-research/bert}}. Our underlying model of choice was the {\tt BERT$_{base,uncased}$} --- which trains a total of 110 million parameters, contains 12 transformer blocks and 12 self-attention heads with a hidden layer dimension of 768 . This model simply draws the [CLS] token embedding of the second to last layer of the BERT model for classification. During the training process the weights of the model are modified slightly to better cater to the task at hand. We finetune hyper parameters such as the learning rate, batch size and maximum sequence length to improve performance for different categories. In this case the model only takes advantage of the data augmentation and preprocessing techniques mentioned above, no extra image and dense features were provided.


\section{Experimental Setup and Results}
\label{sec:expsetup}

The dataset was provided to task participants as a pre-annotated training dataset (containing 7001 samples) and an un-annotated test dataset (containing 1878 samples). We first perform a model and ablation analysis using 10-fold cross validation methods on the training dataset and follow up with scores obtained by our systems on the test dataset, as provided by the task organizers. All results are represented using Accuracy and Macro F$_1$ metrics which were used for ranking the systems by the task organizers.

\begin{table*}[t!]
    \small
     \resizebox{\linewidth}{!}{
    \begin{tabular}{l|c|c|c|c|c|c|c|c|c|c}
         \tabletop
           \multirow{2}{*}{Model} & 
            \multicolumn{2}{c}{Sentiment Analysis} &
            \multicolumn{2}{|c|}{Humor} &
            \multicolumn{2}{c|}{Sarcasm} &
            \multicolumn{2}{c|}{Offensive} &
            \multicolumn{2}{c}{Motivational}\\
     \noalign{\smallskip}
     \cline{2-11}
     \noalign{\smallskip}
        
            & F$_1$ (\%) &
            Acc.(\%) &
            F$_1$(\%) &
            Acc.(\%) &
            F$_1$(\%) &
            Acc.(\%) &
            F$_1$(\%) &
            Acc.(\%)&
            F$_1$(\%) &
            Acc.(\%)\\
        \tablemid
            Logistic Regression &
            {31.97} &
            \textbf{53.84} &
            \textbf{25.85} &
            {31.67} &
            \textbf{23.36} &
            {42.12} &
            \textbf{24.96} &
            \textbf{35.74} &
            {49.71} &
            {53.75} \\
        \tablemid
            Attention based Learner &
            {24.43} &
            {36.67} &
            {22.48} &
            {30.00} &
            {19.62} &
            {25.54} &
            {23.18} &
            {31.18} &
            {46.82} &
            \textbf{60.58} \\
        \tablemid
            Fine-tuned BERT &
            \textbf{32.47} &
            {49.89} &
            {25.20} &
            \textbf{33.00} &
            {22.69} &
            \textbf{43.21} &
            {23.95} &
            {33.58} &
            \textbf{50.29} &
            {56.82} \\
     \tablebot
     \end{tabular}
     }
     \caption{10-fold Cross Validation Results (macro-F$_1$ and accuracy) on subtask A and C, with the training dataset}
     \label{tab:holdout}
\end{table*}

For the training Dataset, Initially, a validation dataset is also maintained to diagnose variance and bias issues that arise in the training phase and to aid hyperparameter tuning. Our overall train, validation, test split ratio is 80:10:10. However, For our final results, we conflate the validation and training set and represent 10-fold cross validation results. We represent our results for Subtask A and the categories within Subtask C (as subtask B and C only vary by semantic levels) in Table \ref{tab:holdout}. We then perform an ablation analysis with a 10-fold cross validation split using the Logistic Regression model to better understand the effect of the various techniques and features employed and how they affect model performance. Due to the task specific nature of the experiment, we decide to carry out the experiment for all the categories --- sentiment analysis, humor, sarcasm, offense and motivation classification. We represent the results in Table \ref{tab:ablation}.
 \ajil{Is the experiments not done in n-fold cross validation ways? n-fold is better because, Cross-validation is usually the preferred method because it gives your model the opportunity to train on multiple train-test splits. is not it?}
\stil{All results replaced with newly measured 10-fold cross validation results}
\ajil{Which Table? Table-2 is Holdout results! You can keep both. is there any much differences?
}
\stil{I forgot to change the table name, sir. There were only minor differences in most cases. There is not enough space to keep both. Also, if we have the 10-fold results, the hold out results are redundant.}

\begin{table*}[t!]
    \centering
    \small
    \resizebox{\linewidth}{!}{
    \begin{tabular}{l|c|c|c|c|c|c|c|c|c|c}
         \tabletop
           \multirow{2}{*}{Features} & 
            \multicolumn{2}{c}{Sentiment Analysis} &
            \multicolumn{2}{|c|}{Humor} &
            \multicolumn{2}{c|}{Sarcasm} &
            \multicolumn{2}{c|}{Offensive} &
            \multicolumn{2}{c}{Motivational}\\
     \noalign{\smallskip}
     \cline{2-11}
     \noalign{\smallskip}
            & F$_1$ (\%) &
            Acc.(\%) &
            F$_1$(\%) &
            Acc.(\%) &
            F$_1$(\%) &
            Acc.(\%) &
            F$_1$(\%) &
            Acc.(\%)&
            F$_1$(\%) &
            Acc.(\%)\\
        \tablemid
            {\tt{\textbf{TFIDF Word (1,2)-gram + Dense Features}}} &
            {27.15} &
            \textbf{59.15} &
            {23.01} &
            {33.84} &
            {17.60} &
            \textbf{49.89} &
            {22.04} &
            {37.57} &
            {42.09} &
            \textbf{64.29} \\
        \tablemid
            \quad + balanced training &
            {30.88} &
            {55.69} &
            {25.99} &
            {31.82} &
            {22.41} &
            {44.07} &
            {24.40} &
            {35.42} &
            {49.23} &
            {53.29} \\
        \tablemid
            \quad + augmentation &
            {28.15} &
            {57.95} &
            {23.81} &
            {33.84} &
            {18.70} &
            {49.04} &
            {22.76} &
            {37.44} &
            {43.87} &
            {63.21} \\    
        \tablemid
            \quad + image features &
            {27.29} &
            {59.08} &
            {23.35} &
            \textbf{33.94} &
            {22.33} &
            {37.84} &
            {22.33} &
            \textbf{37.84} &
            {42.36} &
            {64.18} \\    
         \tablemid
            \quad + balanced training + augmentation &
            {31.51} &
            {54.06} &
            {25.38} &
            {31.45} &
            {23.35} &
            {42.54} &
            {24.95} &
            {35.34} &
            {49.01} &
            {53.26} \\
        \tablemid
            \quad + balanced training + image features &
            {31.25} &
            {55.49} &
            \textbf{26.02} &
            {31.75} &
            {22.72} &
            {43.28} &
            {24.92} &
            {35.82} &
            {49.56} &
            {53.32} \\
        \tablemid
            \quad + augmentation + image features &
            {28.43} &
            {58.16} &
            {23.58} &
            {33.45} &
            {18.78} &
            {48.72} &
            {23.09} &
            {37.84} &
            {44.14} &
            {62.96} \\
         \tablemid
            \quad + balanced training + augmentation + image features &
            \textbf{31.97} &
            {53.84} &
            {25.85} &
            {31.67} &
            \textbf{23.36} &
            {42.12} &
            \textbf{24.96} &
            {35.74} &
            \textbf{49.71} &
            {53.75}  \\
        \tablebot
    \end{tabular}}
    \caption{Ablation study results (macro-F$_1$ and accuracy) on subtask A and C, with the training dataset}
    \label{tab:ablation}
\end{table*}

On the test Datastet, the task organizers rank each system based on averaged Macro F$_1$. For subtasks B ad C where there are 4 separate categories within the task an average score over the four categories was provided. Ranks were provided for only the top submission of each team. Initially, we submit three different systems for task evaluation --- the Logistic Regression model, the BiLSTM + Attention model, the BERT model. However, we saw that the certain models performed better on certain subtasks. Therefore, for our final evaluation a mixed system (referred to in Table \ref{tab:ranks} as Final Mixed System) --- BERT for Subtask A, Attention for Subtask B (except motivational category) and Logistic Regression for Subtask C (including motivational category in Subtask B) --- this represents our official submission to the task. We also represent the macro F$_1$ score and potential ranks of the individual models in Table \ref{tab:ranks}.

\label{sec:results}

\begin{table*}[t!]
    \small
    \centering
    \begin{tabular}{l|c|c|c|c|c|c}
         \tabletop
           \multirow{2}{*}{Model} & 
            \multicolumn{2}{c}{Subtask A} &
            \multicolumn{2}{|c|}{Subtask B} &
            \multicolumn{2}{c}{Subtask C} \\
     \noalign{\smallskip}
     \cline{2-7}
     \noalign{\smallskip}
            & F$_1$ (\%) &
            Rank &
            F$_1$(\%) &
            Rank &
            F$_1$(\%) &
            Rank \\
        \tablemid
            \textbf{Final Mixed System} &
            \textbf{32.48} &
            \textbf{24} &
            \textbf{49.94} &
            \textbf{11} &
            \textbf{30.74} &
            \textbf{15} \\
        \tablemid
            ML Model &
            {31.44} &
            {28} &
            {49.98} &
            {11} &
            {30.74} &
            {15} \\
        \tablemid
            Attention Model &
            {31.68} &
            {28} &
            {49.34} &
            {18} &
            {28.69} &
            {21} \\
        \tablemid
            BERT Model &
            {32.48} &
            {24} &
            {49.75} &
            {11} &
            {29.81} &
            {19} \\
     \tablebot
     \end{tabular}
     \caption{Test Set Results (macro-F$_1$ and potential ranks) on all subtasks. Official submission in bold.}
     \label{tab:ranks}
\end{table*}

Our model study revealed that Logistic Regression and the BERT model exhibit the best results, trading places based on what metric was considered. The results of the Attention based learner were underwhelming and were not competitive with the other models. 

Our ablation analysis reveals that in general balancing and augmentation techniques provide the biggest performance improvements (on the basis of macro F$_1$). We also note that both these techniques were implemented to address the data scarcity and imbalance issue. Image features too, provide slight improvements when added (in all combinations). However, when these techniques are considered in combination, the addition of augmentation over \textit{text + balancing + image features}, lead to a drop in performance for the humor category and also in the sarcasm task, where the \textit{text + image features} model outperforms the \textit{text + augmentation + image features} model. In an additional test, we also observed that for the Attention based learner, image information provided little to no improvement on the performance.

On the test dataset, again BERT and the Logistic Regression model obtain the best results. We also observed that the Attention based learner exhibits lower performance on addition of image features, on all sub tasks.

\section{Error Analysis}
\label{sec:error_analysis}
Due to the under-representation of specific labels, we saw our models were unable to classify samples into these classes effectively. This was especially apparent for the highly under-represented {\tt hateful\_offensive} (of the offensive category) and {\tt very\_twisted} (of the sarcasm category) where very few samples could be correctly classified. While we have reservations regarding the subjective nature of humor and offense, we generally found offensive memes wrongly classified as non-offensive and funny memes wrongly classified as not funny when the context mostly derived from the image. This points to our inability to capture image information effectively. This also indicates the absence of any background knowledge, which is imperative when understanding references that most memes tend to invoke. On analyzing the errors and ground truth of the motivation category, we find many of the annotations puzzling. We are, therefore, more curious as to what the annotation guidelines for this category are and refrain from making any conjectures regarding the classification errors. Another issue we would like to explore is the underwhelming results of the BiLSTM + Attention model. The main issues faced during the training phase are the small overall dataset size and under-representation of specific labels. This led to the inability to train larger and deeper models without facing high variance issues. Another issue that arose due to the small dataset size is the inability to train an image feature extractor from scratch (to alleviate data dissimilarity between the dataset and pretrained models) and better harness image information.
    

\section{Discussion}
\label{sec:discussion}

In our time with the task of meme emotion analysis, we have come to believe that there are many inherent challenges in modelling the task apart from the machine learning algorithms being leveraged. We want to bring attention to the fact that memes are ever-changing --- new memes enter and old memes leave the meme ecosystem very frequently and are quite often informed by current affairs. This could imply that training a model on historical meme samples could provide little insight into the meme ecosystem of today. The trends learnt by training with historical samples might not translate well to the trends of today. In the data collection and annotation phase, we make the following observations:

\begin{figure}[ht]
    \centering
    \includegraphics[scale = 0.27]{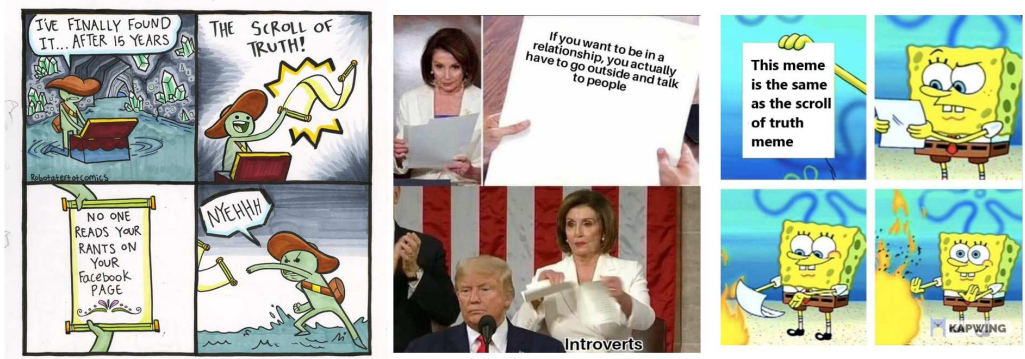}
    \caption{A meme template that can use different referential images}
    \label{fig:memetemplate}
\end{figure}

\textbf{Meme Heterogeneity:} Memes come in different forms and styles --- or so-called templates. These templates can either represent a certain punchline or message and may not share the same referential image. For example, the set of memes in Figure \ref{fig:memetemplate} represent the same idea of `ignoring a fact or truth' but use different referential images to enhance the humor. 
Memes belonging to the same template can be assumed to convey a similar message. There exists a countless number of such templates that are used online. Therefore, we think datasets should not attempt to draw memes from all templates but certain ubiquitous ones --- as in the former case, the dataset does not contain enough samples to adequately represent the meme template's idea. As a product of this, models face the insurmountable task of learning trends from a large number of templates, each with a minimal set of samples. It may be apt to reverse this idea and create datasets with few selected templates but a large number of samples corresponding to each template.

\textbf{Annotator Bias:} As humans, many of us share different tastes and consequently have different senses of humor. What one may find hilarious might not affect another. Therefore, it may be the case that in tasks such as these, we are modelling humor specific to the annotators --- thereby working with an annotator bias. In a recent work by \newcite{WellerSeppi2019}, it was seen that a sample of random humans could not outperform a model trained on the same jokes. This indicates that the model has learned annotator specific biases and may not generalise well to the populace. We are also curious as to what could be the human error on a task such as this.

To address these concerns, we perform an experiment that randomly samples 100 memes from the training dataset (previously annotated) and gets four independent annotators to tag them on all five categories --- sentiment, humor, sarcasm, offensive and motivational. The annotation process was mainly carried out in-house by the authors of this manuscript. We use basic guidelines (in the form of dictionary definitions and examples from the dataset) on tagging the categories and labels. We then calculate macro F$_1$ scores and accuracy for each annotator and compare their performance with the Logistic Regression model's predictions on the same 100 memes. We also calculate inter-annotator agreement metrics to understand better if all the annotators are on the same page. 
We represent the results of this in Table \ref{tab:annotator}. On calculating \newcite{Randolph:2005} free-marginal multi-rater Kappa metrics, we found poor agreement on annotations over all the categories --- Sentiment (0.22), humor (0.10), Sarcasm (0.09), Offense (0.36), Motivational (0.38). The poor agreement scores are an indicator of how concepts such as humor and offense can be subjective. We also find that the human annotators barely outperform our Logistic Regression model, which are indicators of low human performance in such tasks. Few Annotators also report different memes as random or nonsensical. On manual checking, we found these memes to be related to pieces of media (e.g., Star Wars, Lord of the Rings) that the annotator was not familiar with. 

\begin{table*}[t!]
    \small
     \resizebox{\linewidth}{!}{
    \begin{tabular}{l|c|c|c|c|c|c|c|c|c|c}
         \tabletop
           \multirow{2}{*}{Annotators} & 
            \multicolumn{2}{c}{Sentiment Analysis} &
            \multicolumn{2}{|c|}{humor} &
            \multicolumn{2}{c|}{Sarcasm} &
            \multicolumn{2}{c|}{Offensive} &
            \multicolumn{2}{c}{Motivational}\\
     \noalign{\smallskip}
     \cline{2-11}
     \noalign{\smallskip}
            & F$_1$ (\%) &
            Acc.(\%) &
            F$_1$(\%) &
            Acc.(\%) &
            F$_1$(\%) &
            Acc.(\%) &
            F$_1$(\%) &
            Acc.(\%)&
            F$_1$(\%) &
            Acc.(\%)\\
        \tablemid
            Annotator 1 &
            {24.28} &
            {26.00} &
            {20.83} &
            {29.00} &
            {19.08} &
            {30.00} &
            {36.67} &
            {53.00} &
            {53.38} &
            {68.00} \\
        \tablemid
            Annotator 2 &
            {27.92} &
            {32.00} &
            {18.64} &
            {22.00} &
            {25.40} &
            {35.00} &
            {31.82} &
            {41.00} &
            {46.15} &
            {49.00} \\
        \tablemid
            Annotator 3 &
            {33.21} &
            {38.00} &
            {21.69} &
            {30.00} &
            {24.34} &
            {32.00} &
            {24.37} &
            {34.00} &
            {50.47} &
            {66.00} \\
        \tablemid
            Annotator 4 &
            {32.54} &
            {36.00} &
            {28.32} &
            {32.00} &
            {34.63} &
            {39.00} &
            {27.24} &
            {44.00} &
            {47.92} &
            {72.00} \\
        \cline{1-11}
        \tablemid
            \textbf{Annotator Average} &
            {29.49} &
            {33.00} &
            {22.37} &
            {28.25} &
            {25.86} &
            {34.00} &
            {30.02} &
            {43.00} &
            {49.48} &
            {63.75} \\
        \tablemid
            \textbf{Model Performance} &
            {43.79} &
            {59.00} &
            {21.61} &
            {27.00} &
            {22.48} &
            {47.00} &
            {26.37} &
            {38.00} &
            {51.00} &
            {57.00} \\    
     \tablebot
     \end{tabular}
     }
     \caption{Annotator Bias Experiment}
     \label{tab:annotator}
\end{table*}

\section{Conclusion}
\label{sec:conclusion}

Our work on this dataset highlights the difficulty in effectively harnessing image information to understand reference and media that are part and parcel of a meme. Consistent with previous work, we are only able to obtain small performance improvements by employing image-based features. We also highlight modelling issues that arise due to the heterogeneous nature of data (a small number of samples, a large number of templates) and the high variance issues. We also bring to question --- annotation and data collection practices that might not immediately translate to the task of meme analysis. A general suggestion would be to use a large number of annotators or to harness popularity metrics such as likes and upvotes used on websites like Reddit and Instagram. Popularity metrics could provide a more generalized view of the humor or other characteristics of the meme. We think that memes can be imperative to understanding user sentiment in the present internet ecosystem and beyond. We look forward to the new interest that enters this space due to work done here and in the memotion shared task.



\bibliographystyle{coling}
\bibliography{semeval2020-ref}
\end{document}